\documentclass{article}

\usepackage{microtype}
\usepackage{graphicx}
\usepackage{subcaption}
\usepackage{booktabs} 

\usepackage{hyperref}

\usepackage[accepted]{icml2026}

\usepackage{amsmath}
\usepackage{amssymb}
\usepackage{mathtools}
\usepackage{amsthm}
\usepackage{multirow}

\usepackage[capitalize,noabbrev]{cleveref}

\theoremstyle{plain}

\theoremstyle{definition}

\theoremstyle{remark}

\usepackage[textsize=tiny]{todonotes}

\icmltitlerunning{YOCO++: Enhancing YOCO with KV Residual Connections for Efficient LLM Inference}

\begin{document}

\twocolumn[
  \icmltitle{YOCO++: Enhancing YOCO with KV Residual Connections \\
  for Efficient LLM Inference}



  \icmlsetsymbol{equal}{*}

  \begin{icmlauthorlist}
  \icmlauthor{You Wu}{s,a}
  \icmlauthor{Ziheng Chen}{a}
  \icmlauthor{Yizhen Zhang}{a}
  \icmlauthor{Haoyi Wu}{s,a}
  \icmlauthor{Chengting Yu}{z,a}
  \icmlauthor{Yuchi Xu}{a}
  \icmlauthor{Wenbo Su}{a}
  \icmlauthor{Bo Zheng}{a}
  \icmlauthor{Kewei Tu}{s}
  \end{icmlauthorlist}

  \icmlaffiliation{a}{Alibaba Group}
  \icmlaffiliation{s}{ShanghaiTech University}
  \icmlaffiliation{z}{Zhejiang University}


  \vskip 0.3in
]



\printAffiliationsAndNotice{}  

\begin{abstract}
Cross-layer key-value (KV) compression has been found to be effective in efficient inference of large language models (LLMs).
Although they reduce the memory consumption of the KV cache, such methods usually introduce non-negligible performance degradation.
In this work, we aim to enhance the performance of YOCO, a cross-layer KV compression method that shares the KVs of the middle layer with the top-half layers.
We propose YOCO++, an enhanced YOCO that incorporates a weighted residual connection between the KVs of each bottom-half layer and the bottom layer.
Compared to YOCO, YOCO++ increases model capacity while maintaining the same training and inference efficiency.
Our experiments show that YOCO++ achieves state-of-the-art performance among the cross-layer KV compression methods at a 50\% KV cache compression rate, outperforming the standard Transformer.
\end{abstract}

\section{Introduction}

The key-value (KV) cache is a major bottleneck for the inference of large language models (LLMs) \cite{kwon2023efficient} based on the Transformer \cite{vaswani2017attention}.
Various KV cache compression methods \cite{ainslie2023gqa, zhang2023h2o} have been proposed, among which cross-layer compression \cite{wu2024layer, sun2024you, brandon2024reducing} is a widely studied category.
These methods reduce the memory consumption of KV cache by retaining the KVs of only a subset of Transformer layers and reusing them in the remaining layers.
Mainstream cross-layer compression methods directly share the KVs of a specific layer with other layers.
For example, You Only Cache Once (YOCO) \cite{sun2024you} shares the KVs of the middle layer with the top-half layers.

Although the cross-layer KV sharing methods improve inference efficiency, they usually introduce non-negligible performance degradation.
The main cause of this degradation is that the direct KV sharing from one layer to another weakens the model capacity.
To alleviate this problem, one solution is to use a weighted combination of the KVs of multiple layers in a specific layer.
FusedKV \cite{lin2025reconstructing} applies this mechanism to YOCO by using the weighted combination of the KVs of the bottom and middle layers in the top-half layers, achieving a significant performance improvement but also introducing additional cache I/O overhead.

In this work, we further explore the KV combination mechanism by using weighted combinations of the KVs of the current layer and other layers in layers that compute their own KVs, which can be viewed as incorporating residual connections between KVs \cite{zhou2025value}.
We propose YOCO++, which enhances YOCO by introducing a weighted residual connection between the KVs of each bottom-half layer and the bottom layer.
To avoid additional cache I/O overhead during decoding, we cache the combined KVs instead of the original ones.
Our experiments show that YOCO++ maintains the same inference efficiency as YOCO, and achieves state-of-the-art performance cross-layer KV compression methods at a 50\% KV cache compression rate.
Our code is available at \url{https://github.com/wuyou2002/YOCO-plus}.



\section{Related Work}

\subsection{Cross-Layer KV Cache Compression}

Cross-layer KV cache compression methods retain the KVs of a subset of Transformer layers and reuse them in the other layers.
Layer-Condensed KV Cache (LCKV) \cite{wu2024layer} shares the KVs of the top layer with the other layers.
You Only Cache Once (YOCO) \cite{sun2024you} shares the KVs of the middle layer with the top-half layers.
Cross-Layer Attention (CLA) \cite{brandon2024reducing} shares the KVs of the lower layer with the upper layer in each pair of adjacent layers.
A systematic study \cite{wu2025systematic} is conducted among different cross-layer KV sharing methods, including LCKV, YOCO and CLA.
The study finds an effective new configuration called sandwich-middle, which shares the KVs of the middle layer with the other layers.
SVFormer \cite{zhou2025value} shares the values of the bottom layer with the other layers.
SkipV1Former \cite{wu2025improving} shares half of the value heads of the bottom layer with the other layers.
FusedKV \cite{lin2025reconstructing} uses weighted combinations of the KVs of the bottom and middle layers in the top-half layers, and FusedKV-Lite shares the keys of the middle layer and the values of the bottom layer with the top-half layers.

\subsection{Enhanced Residual Connections in Transformer}

As an important component of the Transformer, residual connections \cite{he2016deep} establish an information flow between the hidden states of different layers.
In order to enhance the cross-layer information flow in the Transformer, DenseFormer \cite{pagliardini2024denseformer} introduces the Depth-Weighted-Averaging mechanism, which computes a weighted combination of current and past hidden states.
MUDDFormer \cite{xiao2025muddformer} extends DenseFormer by changing the static, learnable residual weights into dynamic ones computed from the hidden states, and using a multiway design to decouple the four input streams of a Transformer layer, which are the query, key, value and residual.
Attention Residuals \cite{team2026attention} computes the dynamic residual weights using a softmax attention over preceding layer outputs, and introduce a block-level variant to reduce the memory footprint while maintaining performance.
Hyper-Connections (HC) \cite{zhu2025hyperconnections} expands residual connections in the width rather than the depth dimension by maintaining multiple residual streams, yielding performance gains but also violating the identity mapping property of residual connection and introduceing additional memory access overhead.
Manifold-Constrained Hyper-Connections (mHC) \cite{xie2025mhc} projects the residual connection space of HC onto a specific manifold to restore the identity mapping property, while incorporating rigorous infrastructure optimization to ensure efficiency.
ResFormer \cite{zhou2025value} enhances the propagation of initial information by introducing value residual connections in addition to the standard hidden residual connections.

\section{Method}

\subsection{Preliminary: YOCO}

YOCO \cite{sun2024you} is a decoder-decoder model with $L$ layers consists of two parts, which is shown in Figure \ref{fig:method} (a).
The bottom-half $\frac{L}{2}$ layers are called the self-decoder, where the middle layer generates the global KV cache.
The top-half $\frac{L}{2}$ layers are called the cross-decoder, which reuses the global KV cache and does not generate KV cache by itself.
The original YOCO uses efficient attention in the self-decoder to achieve higher efficiency, while we use standard attention in this work for fair comparison.
The size of the KV cache is reduced by 50\% because we only need to cache the KVs in the bottom-half layers.
The prefilling latency is also reduced by 50\% because we can exit early before computing the top-half layers.
However, the cross-layer KV sharing mechanism introduce non-negligible performance degradation.

\subsection{YOCO++}

To enhance the performance of YOCO, we introduce a cross-layer KV combination mechanism, which uses a weighted combination of KVs of multiple layers in a specific layer.
If we apply the combination mechanism to the top-half layers, the cache I/O of each layer increases proportionally with the number of KV layers it uses.
To avoid this additional I/O overhead, we instead apply the combination mechanism to the bottom-half standard Transformer layers, which can be viewed as incorporating residual connections between KV.

The question is, which layers' KVs should we add to the KVs of each layer?
In theory, adding the KVs of all previous layers to those of each layer would provide the strongest capacity.
In practice, \citet{zhou2025value} finds that adding only the values of the bottom layer to those of each layer achieves a similar effect to adding the values of all previous layers.
Based on this observation, we add only the KVs of the bottom layer to those of each layer.

\begin{figure}[tb]
    \centering
    \begin{subfigure}{0.45\columnwidth}
        \centering
        \includegraphics[trim = 90mm 20mm 90mm 20mm, clip, width=\columnwidth]{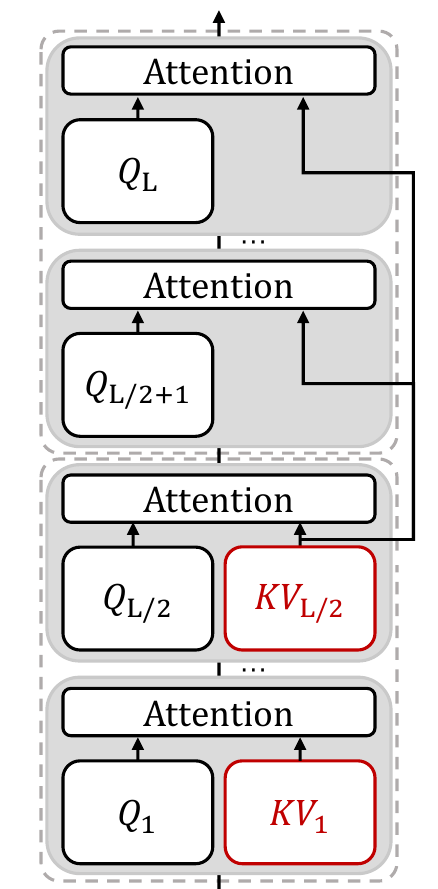}
        \caption{YOCO}
        \label{fig:yoco}
    \end{subfigure}
    \hfill
    \begin{subfigure}{0.45\columnwidth}
        \centering
        \includegraphics[trim = 90mm 20mm 90mm 20mm, clip, width=\columnwidth]{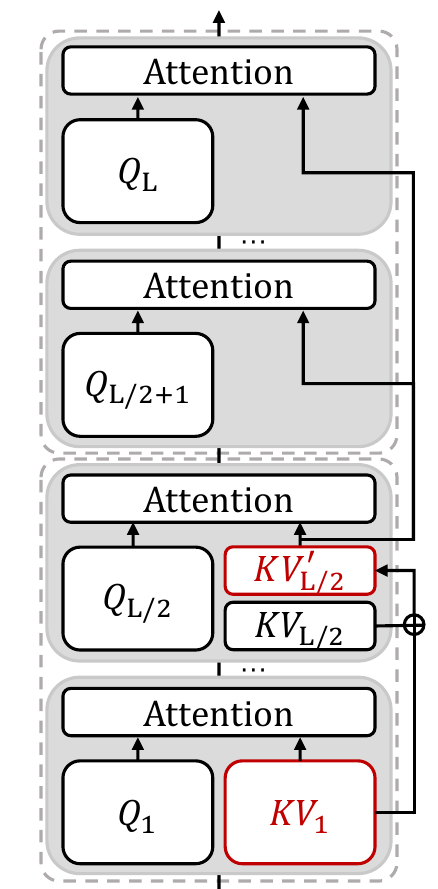}
        \caption{YOCO++}
        \label{fig:yocopp}
    \end{subfigure}
    \caption{Illustration of YOCO and YOCO++. KVs that need to be cached are indicated by red boxes. (a) YOCO caches the KVs of the bottom-half layers and shares the KVs of the middle layer with the top-half layers. (b) Based on YOCO, YOCO++ introduces a weighted residual connection between the KVs of each bottom-half layer and the bottom layer and caches the combined KVs.}
    \label{fig:method}
\end{figure}

By introducing KV residual connections in the bottom-half layers of YOCO, we propose YOCO++.
As shown in Figure \ref{fig:method} (b), in each bottom-half layer $i \in \{2, ..., \frac{L}{2}\}$, we cache the weighted combination of KVs of the current layer $K_i, V_i$ and the bottom layer $K_1, V_1$:
\begin{equation}
    \begin{aligned}
    K_i' &= \alpha^{K}_{i,1} K_1 + \alpha^{K}_{i,2} K_i \\
    V_i' &= \alpha^{V}_{i,1} V_1 + \alpha^{V}_{i,2} V_i
    \end{aligned}
\end{equation}
where $\alpha^{K}_{i,1}, \alpha^{K}_{i,2}, \alpha^{V}_{i,1}, \alpha^{V}_{i,2} \in \mathbb{R}$ are learnable residual weights of layer $i$.
The top-half layers then reuse the cached KVs of the middle layer $K_\frac{L}{2}', V_\frac{L}{2}'$, which carry more information than the original $K_\frac{L}{2}, V_\frac{L}{2}$.
We initialize the residual weights by $\alpha^{K}_{i,1} = \alpha^{V}_{i,1} = 0, \alpha^{K}_{i,2} = \alpha^{V}_{i,2} = 1$ so that the initial state of the model is the same as that of YOCO.

\begin{figure}[tb]
    \centering
    \begin{subfigure}{0.45\columnwidth}
        \centering
        \includegraphics[clip, width=\columnwidth]{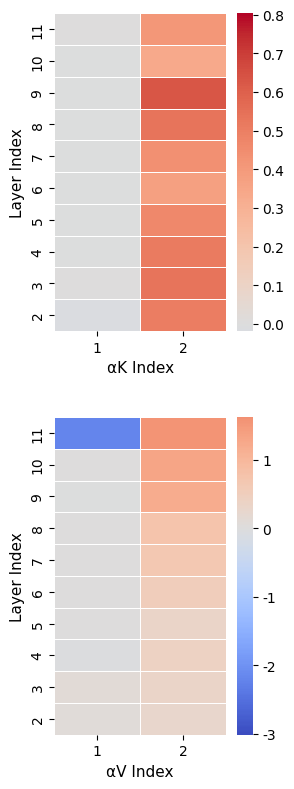}
        \label{fig:alpha1}
    \end{subfigure}
    \hfill
    \begin{subfigure}{0.45\columnwidth}
        \centering
        \includegraphics[clip, width=\columnwidth]{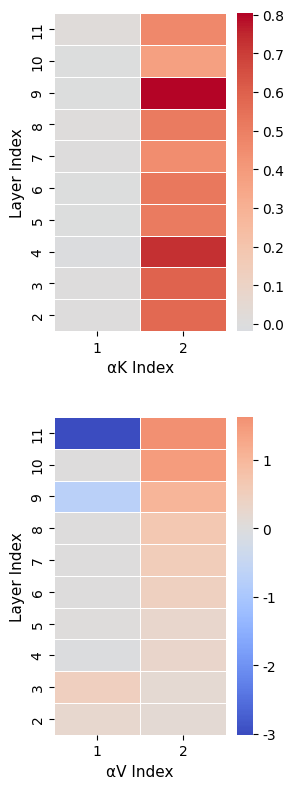}
        \label{fig:alpha2}
    \end{subfigure}
    \caption{Learned residual weights of a 22-layer YOCO++ model without (left) and with (right) the scaling factor.}
    \label{fig:alpha}
\end{figure}

We train YOCO++ following the configuration of TinyLlama \cite{zhang2024tinyllama}, which is a Llama 2 \cite{touvron2023llama} model with 1.1B parameters and $L=22$ layers.
We visualize the learned residual weights $\alpha^{K}_{i},\alpha^{V}_{i}$ for $i \in \{2, ..., 11\}$ in Figure \ref{fig:alpha} (left).
It is shown that $\alpha^{K}_{i,1}, \alpha^{V}_{i,1}$ are close to the initialize value except for $\alpha^{V}_{11,1}$, indicating that these parameters have not been learned well.

In order to facilitate the learning of residual weights, we introduce a scaling factor $\lambda \in \mathbb{R}$, which is a fixed hyperparameter set to be greater than $1$.
The KV residual connections with the scaling factor becomes:
\begin{equation}
    \begin{aligned}
    K_i &= \lambda(\alpha^{K}_{i,1} K_1 + \alpha^{K}_{i,2} K_i) \\
    V_i &= \lambda(\alpha^{V}_{i,1} V_1 + \alpha^{V}_{i,2} V_i)
    \end{aligned}
\end{equation}
The residual weights are now initialized as $\alpha^{K}_{i,1} = \alpha^{V}_{i,1} = 0, \alpha^{K}_{i,2} = \alpha^{V}_{i,2} = \frac{1}{\lambda}$ to keep the initial state unchanged.
We empirically finds that setting $\lambda = 35$ yields the best performance.
After introducing the scaling factor, the learned residual weights differs from the initialize values more obviously, as shown in Figure \ref{fig:alpha} (right).
While $|\alpha^{V}_{i,1}| > |\alpha^{V}_{i,2}|$ in several layers, $\alpha^{K}_{i,1}$ is still close to $0$ in all layers, which is consistent with the asymmetry of keys and values indicated by \citet{lin2025reconstructing}.

\subsection{Efficiency Analysis}

We emphasize that YOCO++ has the same efficiency as YOCO in both training and inference.
In terms of model parameters, YOCO++ only adds four parameters $\alpha^{K}_{i,1}, \alpha^{K}_{i,2}, \alpha^{V}_{i,1}, \alpha^{V}_{i,2}$ in each layer $i \in \{2, ..., \frac{L}{2}\}$ to YOCO, which is negligible.
In terms of computation, YOCO++ only additionally performs a weighted combination on top of YOCO between $K_1, V_1$ and $K_i, V_i$ in each layer $i \in \{2, ..., \frac{L}{2}\}$, whose computational overhead is negligible compared to that of the attention mechanism.
In terms of cache memory, YOCO++ caches $K_1, V_1$ and $K'_i, V'_i$ for $i \in \{2, ..., \frac{L}{2}\}$, whose memory consumption is the same as that of YOCO, which caches $K_i, V_i$ for $i \in \{1, ..., \frac{L}{2}\}$.
In terms of cache I/O, during the training and prefilling stages, YOCO++ requires accessing $K_1, V_1$ and $K_i, V_i$ in each layer $i \in \{2, ..., \frac{L}{2}\}$, introducing additional I/O overhead compared to YOCO.
However, since the efficiency of the training and prefilling stages is bounded by computational overhead, it will not be affected by the additional I/O overhead.
During the decoding stage, YOCO++ requires accessing the cached $K'_i, V'_i$ of the previous tokens, and the newly computed $k_1, v_1$ and $k'_i, v'_i$ of the current token in each layer $i \in \{2, ..., \frac{L}{2}\}$.
Since the additional I/O overhead of accessing the current KVs is negligible compared to that of accessing the previous KVs, the total I/O overhead of YOCO++ the same as that of YOCO.

\section{Experiments}

We conduct experiments to compare the efficiency and performance of YOCO++ with the standard Transformer, YOCO \cite{sun2024you}, FusedKV and FusedKV-Lite \cite{lin2025reconstructing}.
The models follow the configuration of TinyLlama \cite{zhang2024tinyllama}, a 1.1B model built on the architecture and tokenizer of Llama 2 \cite{touvron2023llama}.
The models have 22 layers, 32 attention heads and 4 KV heads.
Our implementation is based on HuggingFace Transformers \cite{wolf2020transformers} with FlashAttention 2 \cite{dao2024flashattention}.

\begin{figure*}[tbh]
    \centering
    {\includegraphics[width=\linewidth]{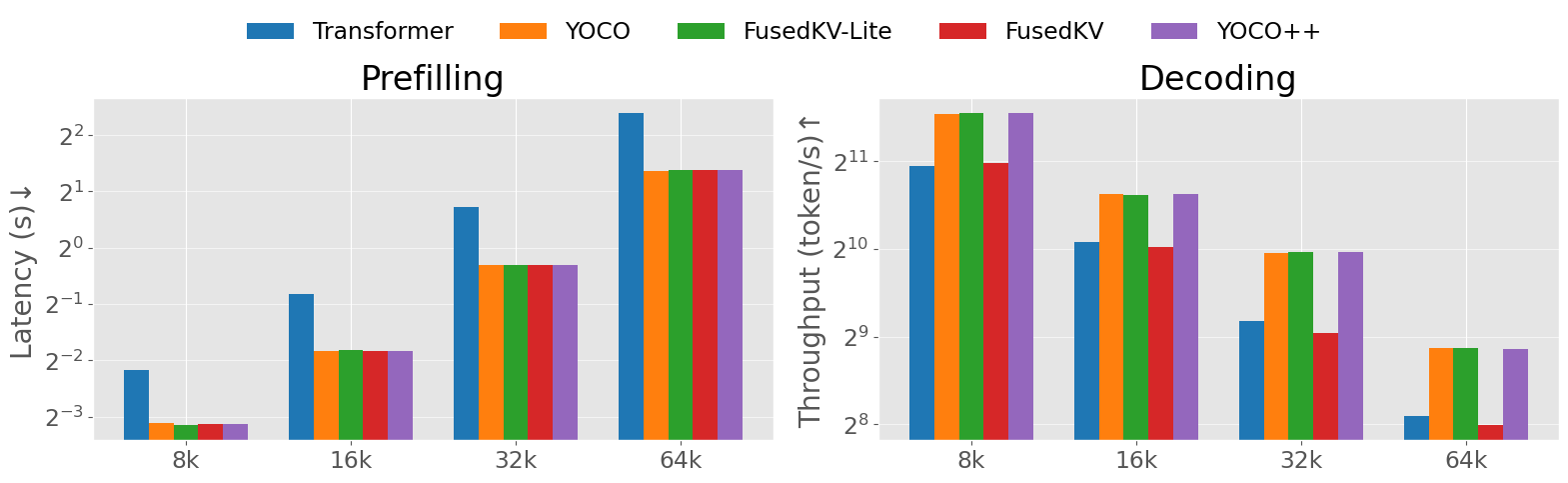}}
    \caption{
      Prefilling latency (left) and decoding throughput (right) on an NVIDIA H20 (96GB) GPU at different sequence lengths.
    }
    \label{figs:throughput}
\end{figure*}

\begin{table*}[tbh]
\centering
\begin{tabular}{@{}lccccccc@{}}
\toprule
Model & HellaSwag & OBQA & WinoGrande & ARC-c & ARC-e & PIQA & Avg \\
\midrule
Transformer & 51.25 & 33.0 & \textbf{56.04} & 27.56 & 52.1 & 70.29 & 48.37 \\
YOCO & 51.21 & 32.0 & 53.59 & \textbf{29.01} & 51.43 & 70.62 & 47.98 \\
FusedKV-Lite & 52.28 & 32.8 & 54.93 & 28.24 & 52.19 & 70.13 & 48.43 \\
FusedKV & 52.04 & 32.0 & 55.48 & 28.67 & 51.98 & 70.35 & 48.42 \\
YOCO++ & \textbf{52.30} & \textbf{34.0} & 55.48 & 27.90 & \textbf{53.11} & \textbf{71.16} & \textbf{48.99} \\
\bottomrule
\end{tabular}
\caption{Zero-shot accuracy on commonsense reasoning tasks.}
\label{tab:performance}
\end{table*}

\begin{figure}[tbh]
    \centering
    {\includegraphics[width=\columnwidth]{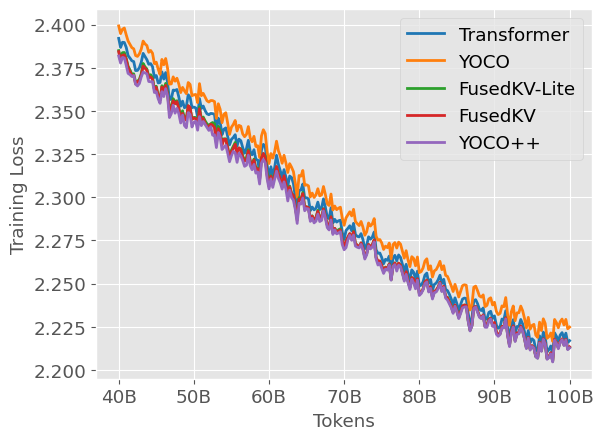}}
    \caption{
      Training loss curves on a 100B subset of the SlimPajama dataset.
    }
    \label{figs:loss}
\end{figure}

\subsection{Inference Efficiency}

We report the prefilling latency (time to first token, TTFT) and the maximum decoding throughput at different batch sizes of all the models on an NVIDIA H20 (96GB) GPU in Figure \ref{figs:throughput}.
As shown in Figure \ref{figs:throughput} (left), YOCO, FusedKV, FusedKV-Lite and YOCO++ have almost the same prefilling latency, which is half that of the standard Transformer, because they can exit early before computing the top-half layers.
As shown in Figure \ref{figs:throughput} (right), YOCO, FusedKV-Lite and YOCO++ have almost the same decoding throughput, which is consistently higher than that of the standard Transformer at different context lengths, because they reduce the memory consumption, allowing for larger batch sizes.
Although FusedKV has the same memory consumption as YOCO, FusedKV-Lite and YOCO++, its throughput degrades to the same level as the standard Transformer because it introduces additional I/O overhead.


\subsection{Model Performance}
\label{sec:performance}

We train all the models on a 100B subset of the SlimPajama dataset \cite{cerebras2023slimpajama} from scratch.
We use the AdamW optimizer \cite{loshchilov2018decoupled} with $\beta_1 = 0.9, \beta_2 = 0.95$. We use a cosine learning rate schedule with a maximum learning rate of $3 \times 10^{-4}$ and a warmup ratio of 0.015. The final learning rate is $3 \times 10^{-5}$. We use a weight decay of 0.1 and gradient clipping of 1.0. The models are trained on 32 NVIDIA H800 (80G) GPUs. The batch size is 512K tokens.
We test the zero-shot performance on commonsense reasoning tasks including Hellaswag \cite{zellers2019hellaswag}, OpenBookQA \cite{mihaylov2018can}, WinoGrande \cite{sakaguchi2021winogrande}, ARC-Easy and ARC-Challenge \cite{clark2018think}, and PIQA \cite{bisk2020piqa}.
The tests are based on the LM Evaluation Harness framework \cite{eval-harness}.

The training loss curves are shown in Table \ref{figs:loss}, and the downstream task accuracies are shown in Table \ref{tab:performance}.
In terms of both training loss and downstream task accuracy, YOCO underperforms the standard Transformer, while FusedKV-Lite, FusedKV and YOCO++ outperforms the standard Transformer.
Among all the models, YOCO++ achieves the best performance.

\subsection{Ablation Study}

We conduct ablation studies on models trained on a 10B subset of the SlimPajama dataset \cite{cerebras2023slimpajama}.
We use 8 NVIDIA H800 (80G) GPUs and a batch size of 128K tokens.
The other training settings are the same as in Section \ref{sec:performance}.
We evaluate the models on a 10M subset of the SlimPajama validation set and report validation loss in Table \ref{tab:ablation}.
The validation loss of YOCO++ without the scaling factor $\lambda$ is similar to that of YOCO, showing that the residual weights cannot be learned effectively without scaling.
Removing residual connections of keys from YOCO++ also leads to performance degradation, showing that residual connections of both keys and values can improve model performance.

\begin{table}[tbh]
\centering
\begin{tabular}{@{}lc@{}}
\toprule
Model & Valid Loss \\
\midrule
Transformer & 2.4266 \\
YOCO & 2.4372 \\
FusedKV-Lite & 2.4258 \\
FusedKV & 2.4204 \\
\midrule
YOCO++ & \textbf{2.4194} \\
\quad w/o scaling & 2.4333 \\
\quad w/o key residual & 2.4264 \\
\bottomrule
\end{tabular}
\caption{Ablations of YOCO++ trained on a 10B subset of the SlimPajama dataset. We report the validation loss on a 10M subset of the SlimPajama validation set.}
\label{tab:ablation}
\end{table}

\section{Conclusion}

In this work, we propose YOCO++, an enhanced YOCO that incorporates a weighted residual connection between the KVs of each bottom-half layer and the bottom layer.
We introduce a scaling factor to facilitate the learning of residual weights.
Our experiments show that YOCO++ maintains the same inference efficiency as YOCO, and achieves state-of-the-art performance among the cross-layer KV compression methods at a 50\% KV cache compression rate, outperforming the standard Transformer.

\nocite{langley00}

\bibliography{example_paper}
\bibliographystyle{icml2026}




\end{document}